\documentclass[runningheads]{llncs}
\usepackage{graphicx}
\usepackage{float}
\usepackage{caption}
\usepackage{subcaption}
\usepackage{tikz}
\usepackage{microtype}
\usepackage{float}
\usepackage{caption} 
\usepackage{array}
\usepackage{multirow}
\usepackage{url}
\usepackage{amsmath}
\usepackage{MnSymbol}
\usepackage{wasysym}

\newcolumntype{P}[1]{>{\centering\arraybackslash}p{#1}}

\begin{document}
\title{Automated classification of pre-defined movement patterns: A comparison between GNSS and UWB technology}
\titlerunning{Classification of movement patterns collected with GNSS and UWB}
%
\author{
Rodi Laanen\inst{1}\orcidID{0000-0003-3544-0914} \and
Maedeh Nasri\inst{2}\orcidID{0000-0002-8547-0946} \and
Richard van Dijk\inst{1}\orcidID{0000-0002-5796-7883} \and
Mitra Baratchi\inst{1}\orcidID{0000-0002-1279-9310} \and
Alexander Koutamanis\inst{3}\orcidID{0000-0002-0355-1276} \and
Carolien Rieffe\inst{2,4}\orcidID{0000-0002-7584-6698} }
\authorrunning{R. Laanen et al.}
%
\institute{
Leiden Institute of Advanced Computer Science, Leiden University, Leiden, The Netherlands 
\and
Unit of Developmental and Educational Psychology, Institute of Psychology, Leiden University, Leiden, The Netherlands
\and
Department of Management in the Built environment, Delft University of Technology, Delft, The Netherlands
\and
Faculty of Electrical Engineering, Mathematics and Computer Science, University of Twente, Enschede, The Netherlands
}

\maketitle 
\begin{abstract}
Advanced real-time location systems (RTLS) allow for collecting spatio-temporal data from human movement behaviours. Tracking individuals in small areas such as schoolyards or nursing homes might impose difficulties for RTLS in terms of positioning accuracy. However, to date, few studies have investigated the performance of different localisation systems regarding the classification of human movement patterns in small areas. The current study aims to design and evaluate an automated framework to classify human movement trajectories obtained from two different RTLS: Global Navigation Satellite System (GNSS) and Ultra-wideband (UWB), in areas of approximately 100 square meters. Specifically, we designed a versatile framework which takes GNSS or UWB data as input, extracts features from these data and classifies them according to the annotated spatial patterns. The automated framework contains three choices for applying noise removal: (i) no noise removal, (ii) Savitzky Golay filter on the raw location data or (iii) Savitzky Golay filter on the extracted features, as well as three choices regarding the classification algorithm: Decision Tree (DT), Random Forest (RF) or Support Vector Machine (SVM). We integrated different stages within the framework with the Sequential Model-Based Algorithm Configuration (SMAC) to perform automated hyperparameter optimisation. The best performance is achieved with a pipeline consisting of noise removal applied to the raw location data with an RF model for the GNSS and no noise removal with an SVM model for the UWB. We further demonstrate through statistical analysis that the UWB achieves significantly higher results than the GNSS in classifying movement patterns.

\end{abstract}
\section{Introduction}\label{introduction}
Advanced localisation systems have enabled capturing trajectories of moving entities providing insights in various areas such as understanding periodic behavioural patterns~\cite{baratchi2013recognition}, pedestrian movement dynamics~\cite{PETRE2017309}, and anomaly detection~\cite{shih2016personal}. Using such positioning systems in enclosed spaces helps uncover social dynamics information, for instance by tracking the movement behaviours of children in a schoolyard during recess~\cite{luo2020outdoor,nasri2022novel} or the elderly in nursing homes~\cite{lin2018gps,marins2013extending}. 

Global Navigation Satellite Systems (GNSS) are one of the most recently used localisation systems which demonstrated strong promises in urban computing and transportation~\cite{dabiri2018inferring,etemad2018predicting}. However, a GNSS might not provide the desired positioning accuracy in small areas such as schoolyards and nursing homes. Namely, under ideal circumstances, GNSS devices can achieve a positioning accuracy of approximately 2-3 meters~\cite{dabove2019towards,tomavstik2017horizontal}, and the positioning error can deteriorate up to 20 meters or more due to, for instance, multi-path or ionospheric scintillation~\cite{dabove2019towards,kaplan2017understanding}. To mitigate this problem, several studies~\cite{kuhn2008high,mahfouz2008investigation,mimoune2019evaluation} have demonstrated promising results regarding the positioning accuracy through Ultra-wideband (UWB) technology by reporting positioning accuracies between a 10-15 cm range.

In addition to positioning, GNSS and UWB trajectories can be used to analyse human movement behaviour and unravel complex spatio-temporal patterns. Extracting these patterns can provide useful insights in, for instance, children's use of space and interactions with the schoolyard environment during recess~\cite{nasri2022novel} and the development of personalised geofences for elderly people suffering from Alzheimer's disease~\cite{lin2018gps}. However, to the best of our knowledge, there is no study directly comparing the performance of these two technologies in a movement pattern classification task.

To address this gap, the current paper aims to design an automated framework to classify pre-defined human movement patterns collected with GNSS and UWB technologies and to evaluate their respective classification performances. We will address the following questions: (i) How to design an entire pipeline to collect and annotate the GNSS and UWB spatial-temporal data, pre-process and classify them correctly? (ii) How to select the optimum hyperparameter configurations for the data segmentation, noise removal and classification stages within the pipeline which result in the best performance? (iii) Which technology, GNSS versus UWB, achieves higher results in terms of movement pattern classification in small areas of approximately 100 square meters? Moreover, this paper contains the following contributions:
\begin{enumerate}
 \item[$\bullet$] We collected and introduced a GNSS and a UWB data set both consisting of 104 trajectories based on 4 annotated pre-defined human movement patterns.
 \item[$\bullet$] We designed an automated framework which aims to correctly classify pre-defined movement patterns collected with GNSS and UWB technology. The automated framework contains three options for applying noise removal: (i) no noise removal, (ii) Savitzky Golay filter on the raw location data~\cite{etemad2018predicting} or (iii) Savitzky Golay filter on the extracted features~\cite{dabiri2018inferring} as well as three options regarding the classification model: Decision Tree (DT), Random Forest (RF) or Support Vector Machine (SVM).
 \item[$\bullet$] We performed hyperparameter optimization to automate the full trajectory processing pipeline, including data preparation, noise removal and classification stages for both UWB and GNSS trajectories.
\end{enumerate}
The remainder of this paper is organized as follows: In Section 2, we give a brief overview of related works. Section 3 contains the overview of our method. The details of our experiments and the evaluation of the proposed method are presented in Section 4. Lastly, the conclusion and future research directions are discussed in Section 5.
\section{Related works}
Recent studies compared the positioning accuracy of GPS and UWB data within the sport domain~\cite{bastida2018accuracy,waqar2021analysis}. Waqar et al.~\cite{waqar2021analysis} compared the positioning accuracy from a GPS and a UWB system on a tennis field. Based on their quantitative analysis, the authors concluded that UWB outperforms GPS in terms of localisation accuracy~\cite{waqar2021analysis}. In the study of Bastida et al.~\cite{bastida2018accuracy}, the two technologies were assessed on a soccer field where the UWB system proved to achieve higher localisation accuracy compared to the GPS system~\cite{bastida2018accuracy}. Despite the interesting results found in these two studies on positioning accuracy and performance, they do not directly address movement pattern classification tasks performed through GNSS and UWB systems.

Several studies focused on classification of transportation modes via GPS trajectories by designing a competitive framework with classical machine learning algorithms~\cite{etemad2018predicting}, and one with a Convolutional Neural Network~\cite{dabiri2018inferring}. Another interesting framework which clustered basic patterns via deep representation learning is proposed in the work of Yao et al.~\cite{yao2018learning}. However, the aforementioned three studies~\cite{dabiri2018inferring,etemad2018predicting,yao2018learning} are restricted to GNSS trajectories as input and not adapted to UWB trajectories, and secondly, hyperparameters present within each respective framework are either determined manually or without motivation and do not include automated hyperparameter optimization.
\section{Methodology}\label{methodology}
In this section, we first introduce the general outline of the automated framework (see Figure~\ref{automated_framework_fig}). Next, we discuss the following parts of the automated framework in more detail: raw trajectories, data preparation, features, trajectory features, noise removal and normalization.
\subsection{Automated framework}\label{automated_framework}
\begin{figure}[]
\begin{center}
\includegraphics[width=\textwidth]{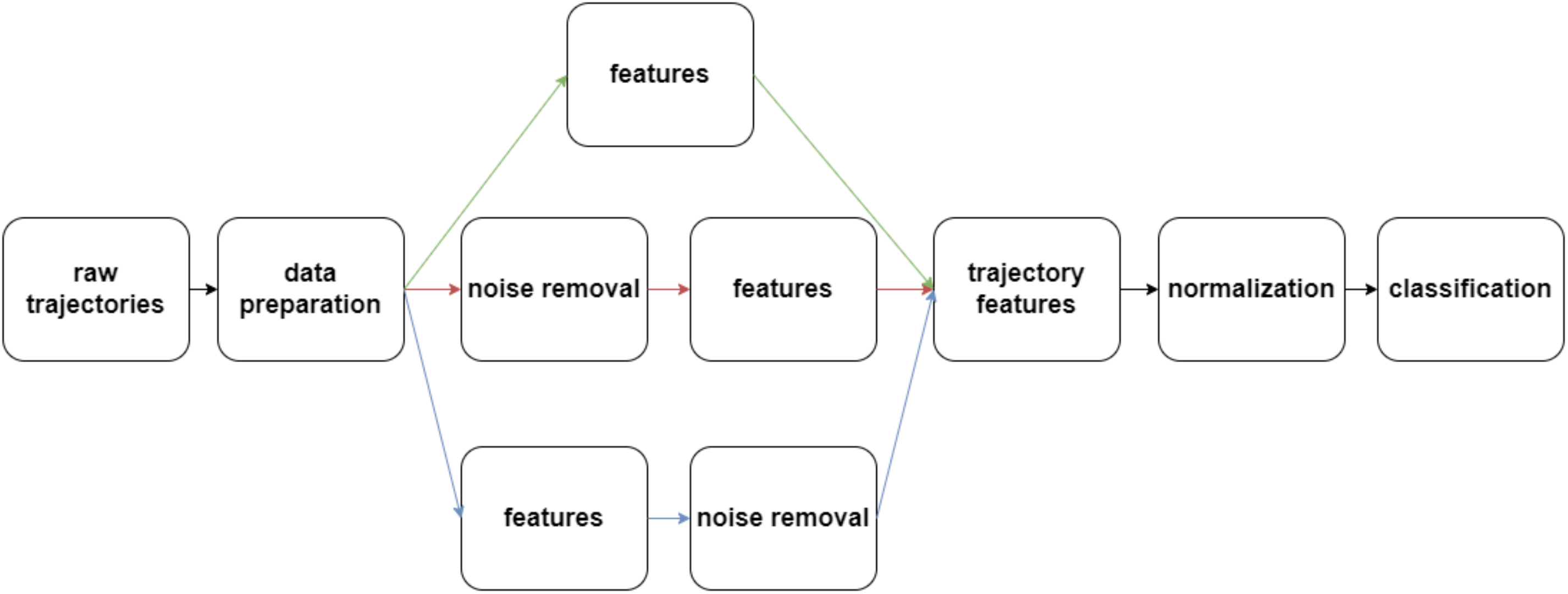}
\end{center}
\caption{Automated framework, inspired by Etemad et al.~\cite{etemad2018predicting}, which takes raw spatial-temporal data from GNSS and UWB systems and ultimately learns from the corresponding annotated movement patterns. Applying no noise removal, the Savitzky Golay filter on the raw location data and the Savitzky Golay filter on the features are indicated by the green, red and blue lines, respectively.}\label{automated_framework_fig}
\end{figure}

The proposed automated framework, depicted in Figure~\ref{automated_framework_fig}, inspired by Etemad et al.~\cite{etemad2018predicting}, automatically takes raw discrete trajectories collected with GNSS or UWB technology as input, pre-processes the raw spatio-temporal data and feeds the obtained data to a classifier. Additionally, we investigated the effect of a noise removal filter, i.e., Savitzky Golay filter which is discussed in Section~\ref{noise_removal_methodology}, on the performance of the entire automated framework by implementing three pipeline configurations in parallel: (i) extracting features without applying noise removal, (ii) applying a Savitzky Golay filter directly on the raw location data~\cite{etemad2018predicting} or (iii) extracting features from the raw spatio-temporal data followed by applying a Savitzky Golay filter on the extracted features~\cite{dabiri2018inferring}. The parallel configurations are denoted by green, red and blue lines in Figure~\ref{automated_framework_fig}, respectively. Furthermore, the automated framework includes three classification algorithms: DT, RF or SVM. In this respect, the automated framework yields nine possible pipeline configurations. 
\newline
\indent As indicated in Figure~\ref{automated_framework_fig}, the automated framework contains 7 distinct steps: 1) raw trajectories, 2) data preparation, 3) features, 4) noise removal, 5) trajectory features, 6) normalization and 7) classification. These steps, except for the classification stage, are explained in detail in the following subsections.
\subsubsection{Raw trajectories:}
Raw trajectories are acquired from localisation systems (i.e., GNSS and UWB) within $R$ timestamps. Each discrete raw trajectory $traj$ contains the spatio-temporal list of points of one moving object as follows:
\begin{equation}\label{rawtrajectory}
traj = \{P_1, P_2, ..., P_R\},~P_{r} = (c1_r, c2_r, t_r), ~r \in[1,R] 
\end{equation}
where $P_r$ defines the 2-D position of the object at timestamp $t_r$. For GNSS trajectories each position is equal to $(c1_r = lat_r, c2_r = lon_r, t_r = t_r)$ 
and for UWB trajectories each position is in Euclidean space as $(c1_r = x_r, c2_r = y_r, t_r = t_r)$. The number of positions R of the GNSS trajectory can be smaller than the number of positions R of the UWB trajectory because of the difference in data rate (1 Hz versus around 6 Hz for UWB).
\subsubsection{Data preparation:}
The data preparation part of the automated framework involves segmenting a raw discrete trajectory $traj$ into $M$ segments with varying size $R_m$. Specifically, a raw discrete trajectory $traj$ is split into $traj\mod M$ segments with size $\lfloor traj / M \rfloor$ + 1 and the remaining segments with size $\lfloor traj / M \rfloor$. Hence, the segments do not overlap time-wise. The whole set of segments covers the trajectory:
\begin{equation}\label{rawtrajectory}
traj = \{traj_1, traj_2, ...., traj_M\}
\end{equation}
\subsubsection{Features:}\label{point_features} From each segment of a GNSS or a UWB trajectory, we extract three features: average velocity, change of velocity and change of angle since these features capture alternations of movement behaviours~\cite{yao2018learning}.
Regarding a GNSS trajectory segment, each record $r \in [1,R_m]$, where $R_m$ refers to the total number of records within this GNSS trajectory segment, contains the GNSS coordinates in ($lat_r, lon_r$), and the corresponding timestamp $t_r$, from which the average velocity is computed as follows:
\begin{equation}\label{haversine}
 v_r = \frac{haversine((lat_{r+1},lon_{r+1}),(lat_r,lon_r))}{t_{r+1} - t_r},~~~v_1 = 0
\end{equation}
where $v$ represents the average velocity, $t_r$ describes the timestamp $r$ $\in$ $[2,R_m]$ indicates a single record of a trajectory segment and $lat_r$ and $lon_r$ denote the latitude and longitude, respectively. By first converting the latitude and longitude coordinates to an Euclidean plane projected on the surface of the earth, the $haversine$ function determines the Euclidean distance between two consecutive GNSS points.
\newline
Additionally, we computed the angle between two subsequent GNSS points by:
\begin{equation}
 X_r = (0, lon_r),~~~
 X_{r+1} = (0, lon_{r+1}),~~~
 Y_r = (lat_r, 0),~~~
 Y_{r+1} = (lat_{r+1}, 0)\\
\end{equation}
\begin{equation}\label{gps_angle}
 a_r = arctan2(haversine(Y_{r+1},Y_r), haversine(X_{r+1},X_r)),~~~a_1 = 0
\end{equation}
where $a_r$ denotes the angle between record $r+1$ and $r$ and $r$ $\in$ $[2,R_m]$.

For each discrete UWB trajectory segment, a single record, $r \in [1,R_m]$, where $R_m$ refers to the total number of records within this UWB trajectory segment $m$, contains the position of an active tag in ($x_r, y_r$) coordinates with corresponding timestamp $t_r$. We computed the average velocity $v_r$ according to the following equations:
\begin{equation}\label{uwb_speed1}
 v_r = \frac{d_r}{t_{r+1}-t_r},~~~v_1 = 0,~~~d_r = \sqrt{({x_{r+1}}-{x_r})^2 + ({y_{r+1}} - {y_r})^2}
\end{equation}
where $x$ and $y$ denote the x and y coordinates, respectively. $d_r$ refers to the Euclidean distance between record $r+1$ and $r$ and $r$ = $[2,R_m]$. Besides, the angle between two consecutive UWB points is calculated as follows:
\begin{equation}
 a_r = arctan2((y_{r+1} - y_r), (x_{r+1} - x_r)),~~~a_1 = 0
\end{equation}

Regarding the three features: average velocity, change of velocity and change of angle, we computed the average velocity of the discrete GNSS trajectory segments with Equation~\ref{haversine} and for the discrete UWB trajectory segments with Equations in~\ref{uwb_speed1}. For both the discrete GNSS and UWB trajectory segments, we computed the change of velocity and change of angle for two subsequent records as follows:
\begin{equation}
 v_{\Delta_r} = v_{r}-v_{r-1},~~~v_{\Delta_1} = 0
\end{equation}
\begin{equation}
 a_{\Delta_r} = a_r-a_{r-1},~~~a_{\Delta_1} = 0
\end{equation}
where $v_\Delta{}_r$ denotes the change of velocity and $a_\Delta{}_r$ represent the change of angle, both for $r$ $\in$ $[2,R_m]$.
\subsubsection{Trajectory features:}
We utilized statistical measures which provide information about the entire trajectory $traj$ or about segments $traj_m$ of the trajectory in question as described by Etemad et al.~\cite{etemad2018predicting}.
In this respect, we selected these features to gain global information about the (segment) trajectories. For the whole (segment) trajectory, 5 global features~\cite{etemad2018predicting} per set of extracted features were derived: the minimum, maximum, mean, median, and standard deviation. Additionally, we computed 5 local features~\cite{etemad2018predicting} per set of extracted features: five percentiles (10, 25, 50, 75, 90). Hence, the global trajectory is extended with 3 extracted features x 5 statistical measures, while a local trajectory segment is extended with 3 extracted features x 5 percentile measures.
\subsubsection{Noise removal:}\label{noise_removal_methodology}
Since both RTLS use electromagnetic waves, the collected trajectories might include positional errors due to e.g., conductors (e.g., metal and water), air humidity, multi-path interference and battery levels~\cite{dabove2019towards,kaplan2017understanding,kuhn2008high,pozyx}. A promising candidate is the Kalman Filter~\cite{barrios2011improving,kalman1960new,kennedy2020improving} that has been used for several purposes such as noise removal. However, according to Kennedy~\cite{kennedy2020improving}, the Kalman Filter can only find an optimal solution if the noise within a signal is normally distributed. Given that the shape of the underlying noise distribution of the raw spatio-temporal data or the extracted features is unknown~\cite{dabiri2018inferring}, the Kalman Filter might not provide optimal results in terms of noise removal. The Savitzky Golay filter mitigates this issue since it does not assume a distribution of the signal's noise~\cite{dabiri2018inferring,kennedy2020improving}. Another advantage is that the pattern of the original signal will be maintained after applying the Savitzky Golay filter~\cite{dabiri2018inferring}.
Therefore, the Savitzky Golay filter~\cite{savitzky1964smoothing} is adopted in the noise removal stage. 

This filter selects a subset of data points around the target point defined by the sliding window size and fits a polynomial on the respective subset~\cite{schafer2011savitzky}. As a final step, the target point gets substituted for the computed outcome of the polynomial~\cite{schafer2011savitzky}. Both the sliding window and the degree of the fitted polynomial are crucial hyperparameters, which will be further discussed in Section~\ref{smac}.
\subsubsection{Normalization:}
Min-max normalization transforms all features into the range $[0,1]$ by mapping the lowest value to 0, the highest value to 1 and the remaining values between 0 and 1. In this way, all generated features will be projected to the same range ($[0,1]$), while the relationship between the original feature set values is preserved~\cite{han2011data}.
\subsection{Automated hyperparameter optimisation}\label{smac}
Previously mentioned studies~\cite{dabiri2018inferring,etemad2018predicting,yao2018learning}, which inspired our design of the automated framework, manually performed hyperparameter Optimisation (HPO). A manual approach demands a thorough understanding regarding the impact of each hyperparameter and can become a time-consuming and inefficient task which might result in less optimal and accurate outcomes~\cite{feurer_hyperparameter_2019}. In the current study, Sequential Model-based Algorithm Configuration (SMAC)
~\cite{lindauer-arxiv21a} is adopted to make the HPO automated for the data preparation, noise removal, and classification stages.

SMAC is a general-purpose HPO library based on the Bayesian optimization framework~\cite{lindauer-arxiv21a}. It allows for optimizing various types of hyperparameters (i.e., categorical, numerical and conditional). When dealing with a full pipeline, flexibility regarding the implementation of hyperparameter types is extremely crucial. Notably, design choices such as the search space benefit from the hyperparameter type flexibility provided by SMAC. Thereby, SMAC includes the option to define conditional hyperparameters which allow for selecting, for instance, the type of machine learning algorithm.
\subsection{Selection of hyperparameters}
To optimize the pipeline, we need to define the search space incorporating all hyperparameters. First, the data preparation stage contains a hyperparameter that specifies the segmentation size of each discrete trajectory $traj_m$. Second, two hyperparameters are present in the noise removal stage: one that determines the respective sliding window-size of the targeted data points and another one which optimizes the degree of the fitted polynomial. Last, in the classification stage, we automatically let the hyperparameters of each respective classifier be optimized. Via SMAC, we aim to achieve higher classification performances assessed with the used score metrics in this study.
Detailed information about the hyperparameters and their respective types and value ranges are described in Table~\ref{table_hyperparameters}.
\section{Experiments}
In this section, first, the data collection sessions under controlled conditions are described. Next, the performance metrics to assess each pipeline configuration and the experimental settings including the bootstrapping protocol are introduced. Finally, we will report and interpret the obtained results in our experiments. All experiments are implemented in Python 3.9.7\footnote{\url{https://www.python.org}} with scikit-learn 1.0.2\footnote{\url{https://scikit-learn.org}}, SciPy 1.7.1\footnote{\url{https://docs.scipy.org}} and SMAC1.3.3\footnote{\url{https://github.com/automl/SMAC3}}. The source code from this study is available on GitHub\footnote{\url{https://github.com/rlaanen/uwbtrajectorypatterns}}.
\subsection{Data sets}
We collected spatio-temporal GNSS data points (i.e., a list of (\textit{longitude}, \textit{latitude}, \textit{timestamp})) with an 1 Hz sampling rate using a Samsung Galaxy Tab A7 (2020) SM-T500 64 GB tablet~\cite{TabA7} and the Sensor Logger software\footnote{\url{https://www.tszheichoi.com/sensorlogger}}. We used the Pozyx Creator Kit~\cite{pozyxcreatorkit} to collect UWB data points (i.e., a list of (\textit{x}, \textit{y}, \textit{timestamp})) with an average sampling rate of 5.91 Hz per active tag. All practical details regarding the GNSS and UWB system configurations and the UWB implementation are described in~\cite{laanen2022classification}.

\begin{figure}[]
\centering
\begin{subfigure}{.5\textwidth}
 \centering
 \includegraphics[width=1\linewidth]{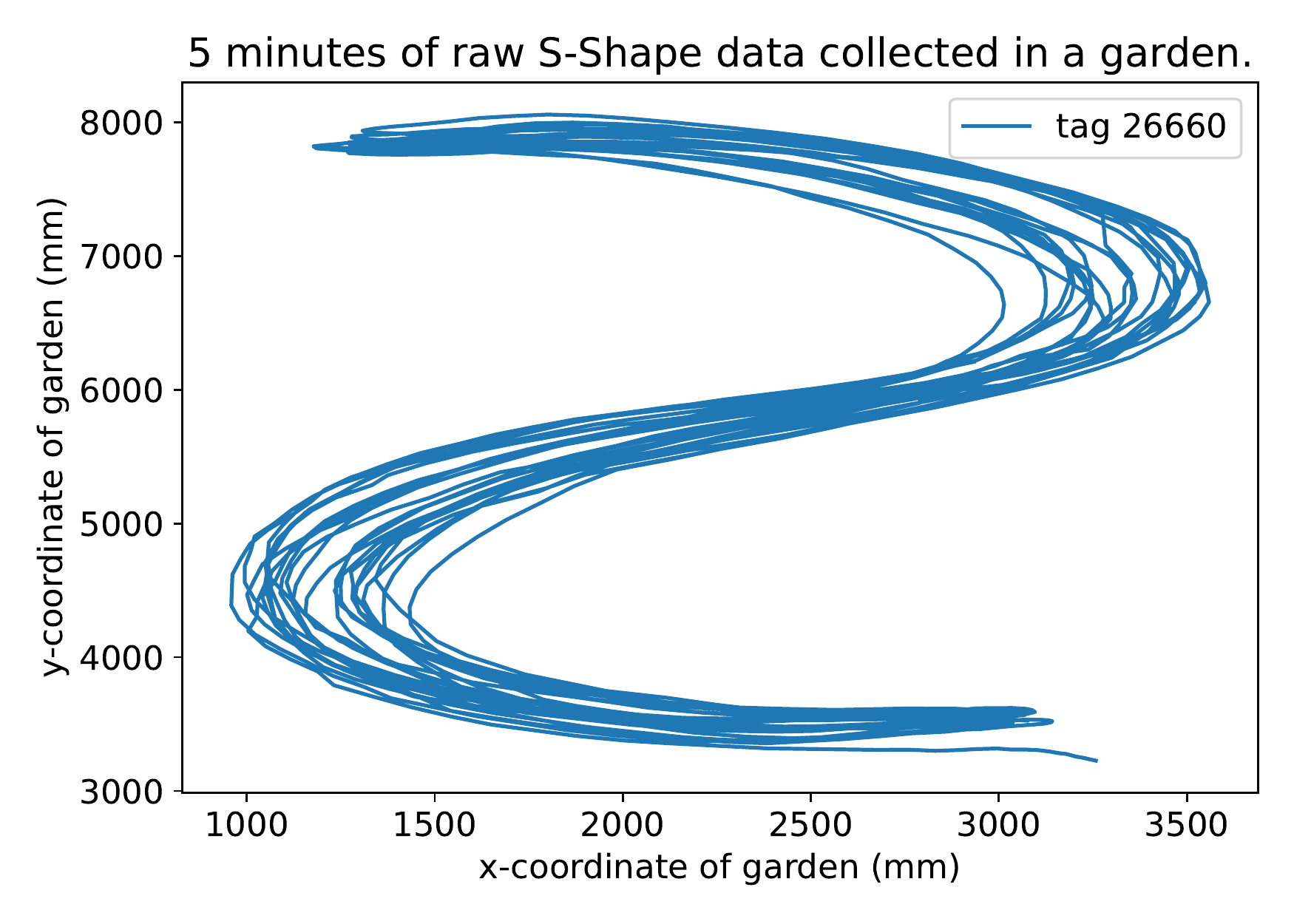}
 \caption{$\textit{S-Shape}$}
 \label{fig:sub1}
\end{subfigure}%
\begin{subfigure}{.5\textwidth}
 \centering
 \includegraphics[width=1\linewidth]{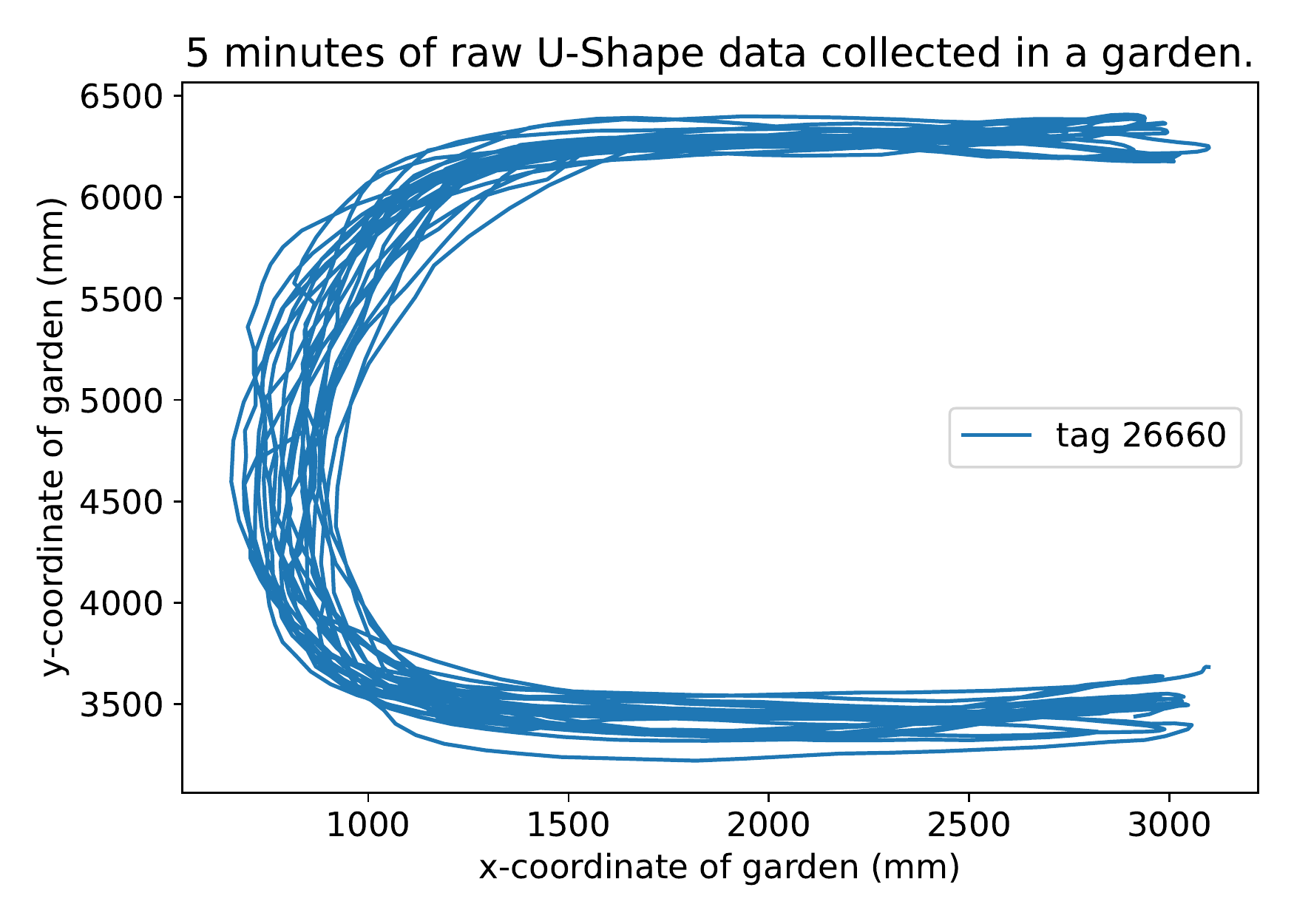}
 \caption{$\textit{U-Shape}$}
 \label{fig:sub2}
\end{subfigure}
\caption{ Visualizations of 5 minutes of (a) $\textit{S-Shape}$ and (b) $\textit{U-Shape}$ patterns collected via UWB technology.}
\label{uwb_shapes}
\end{figure}

During the data collection sessions, we adopted the same movement patterns as Yao et al.~\cite{yao2018learning}, i.e., \textit{Straight} and \textit{Circling} and further extended the \textit{Bending} pattern by defining two new patterns: \textit{S-Shape} and \textit{U-Shape}, as represented in Figure~\ref{uwb_shapes}.
We opted for the \textit{S-Shape} and \textit{U-Shape} as \textit{Bending} patterns since they can be composed of parts of the \textit{Straight} and \textit{Circling} patterns, resulting in four different classes to be recognized, and making the classification task more challenging. 

We collected trajectory data consisting of the movements of four adults who walked a single pattern continuously for 5 minutes by holding a GNSS and a UWB recording device. In this respect, we obtained two data sets each consisting of 19 \textit{Straight}, 25 \textit{circling}, 30 \textit{S-Shape} and 30 \textit{U-Shape} movement patterns, resulting in a total of 104 5-minute-long discrete GNSS and another 104 5-minute-long UWB trajectories.
\subsection{Performance metrics}
Four performance metrics were implemented to assess the performance of the trained model on the test data set. Throughout our experiments, we considered the following four metrics: macro precision, macro recall, macro F$_1$ and Matthew's Correlation Coefficient (MCC). We opted for the macro F$_1$ score since macro-averaging treats all four patterns on an equal basis by being insensitive to class imbalance~\cite{sokolova2009systematic}. Furthermore, the F$_1$ score represents the harmonic mean of the precision and recall. Therefore, we included the macro precision and macro recall score metrics.
Thereby, we adopted an extended version of the MCC metric which applies to multi-class classification tasks~\cite{gorodkin2004comparing}. This score is formulated as follows:
\begin{equation}\label{eq_mccmulti}
 MCC = \frac{c \cdot s - \sum_{k}^{K} p_k \cdot b_k}{\sqrt{({s}^2 - \sum_{k}^{K} {p_k}^2) \cdot ({s}^2 - \sum_{k}^{K} {b_k}^2)}}
\end{equation}
where $c$ represents the total number of correctly predicted instances, $s$ refers to the total number of instances, $K$ is the total number of classes, $k$ refers to a single class, $b_k$ denotes the true occurrences of class $k$ and $p_k$ describes the number of times class $k$ was predicted~\cite{scikit-learn}. This equation explicitly reflects that a high-performance score is correlated with the number of correctly predicted instances~\cite{grandini2020metrics}. Additionally, the MCC is reported to be robust against potential performance bias from an imbalanced data set~\cite{chicco2020advantages}.
\subsection{Experiment settings}\label{experiment_settings}
As discussed in Section~\ref{methodology}, the data preparation, noise removal and classifier steps within the automated framework are controlled by hyperparameters. We decided to automate the HPO process with SMAC~\cite{lindauer-arxiv21a}.
All hyperparameters including their respective value range and type are represented in Table~\ref{table_hyperparameters}.
\begin{table}[]
\caption{hyperparameters including their respective value ranges and types influencing the data preparation, noise removal and classifier steps within the automated framework. Given that the Savitzky Golay filter imposes an odd-sized window, we opted for a Categorical type of hyperparameter.} \label{table_hyperparameters}
\begin{center}
\scalebox{0.8}{
\begin{tabular}{p{4.0cm} | p{3.5cm} | p{4cm} | p{3cm}}
 $\textbf{Method}$ & $\textbf{Hyperparameter}$ &
 $\textbf{Value Range}$ & $\textbf{Type}$ \\
 \hline
 Partitioning Instances & split & [1, 10] & Uniform Integer \\
 \hline
 Savitzky Golay filter & window\_length & [1, 3, ..., 27, 29] & Categorical \\
 & polyorder & [1, 10] & Uniform Integer \\
 \hline
 Decision Tree & max\_depth & [5, 50] & Uniform Integer \\
 & min\_samples\_leaf & [1, 10] & Uniform Integer \\
 & min\_samples\_split & [2, 10] & Uniform Integer \\
 & criterion & ["gini", "entropy"] & Categorical \\
 \hline
 Random Forest & n\_estimators & [5, 100] & Uniform Integer \\
 & min\_samples\_leaf & [1, 10] & Uniform Integer \\
 & max\_depth & [5, 50] & Uniform Integer \\
 & min\_samples\_split & [2, 10] & Uniform Integer \\
 & criterion & ["gini", "entropy"] & Categorical\\
 \hline
 Support Vector Machine & C & [0.1, 100] & Uniform Float \\
 & kernel & ["linear", "poly", "rbf", "sigmoid"] & Categorical
\end{tabular}
}
\end{center}
\end{table}

Although SMAC automates the HPO, it remains a time-consuming task~\cite{feurer_hyperparameter_2019} and setting no random-seed in the SMAC framework makes the optimisation non-deterministic~\cite{lindauer-arxiv21a}. For these reasons, we first assessed the relationship between the wallclock-time and the MCC score in order to obtain the optimum wallclock-time. Second, we implemented an adaptation of the bootstrapping protocol proposed in the work of Wang et al.~\cite{wang2019automated} to obtain sufficient and fair distributions of results and apply a significance test afterwards.

The wallclock-time represents the time in seconds between starting the SMAC optimisation and terminating it. In order to find the wallclock-time SMAC needs to find near-optimal hyperparameter values, we proceeded as follows: First, we split the entire data set into 67\% training and 33\% test set, and then, with increment-steps of 25 seconds, we performed 15 independent optimisation runs per wallclock-time on the pipeline configuration yielding the most hyperparameters (i.e., noise removal with an RF model). This method resulted in a wallclock-time of 500 seconds at which the highest performance reflected in the MCC score was reached. This optimal value is adopted in the bootstrapping protocol.\newline
\indent A single pipeline optimization within the bootstrapping protocol~\cite{wang2019automated} consists of the following steps: First, we split the entire data set again into a 67\% training and 33\% test set after which we exclusively utilise the training set to perform the optimisation with SMAC. For the validation method, we opted for a stratified 10-fold cross-validation with an MCC score metric since k-fold cross-validation segments the training data set in folds, each containing approximately the same distribution of class labels as in the training data set~\cite{scikit-learn}. Given that SMAC targets HPO as a minimisation problem~\cite{lindauer-arxiv21a}, we aggregated over the 10 resulting performances and subtracted the outcome from 1 as an indicator of the respective pipeline configuration performance.\newline
\indent In this respect, we let SMAC optimise a single pipeline configuration 15 times on the training data set. Next, from the 15 generated ones, we randomly sample 5 incumbents from which we select the one yielding the lowest performance score. We repeated this process 50 times for every pipeline configuration. Subsequently, we retrained 50 models according to the 50 best-found incumbents on the training data set. Finally, we evaluated the 50 models on the independent test set and averaged over the 50 scores.
\subsection{Results}
In this section, we first describe the GNSS and UWB results separately. Subsequently, we compare the two technologies according to the pipelines which achieved the highest performance scores.
\subsubsection{GNSS results:}\label{gps_results}
Via the above-mentioned bootstrapping protocol in Section~\ref{methodology}, we optimised the nine possible pipeline configurations on the GNSS data set. The averaged scores on the test data set reflected in macro precision, macro recall, macro F$_1$ and MCC which are based on the 50 best-found hyperparameter settings according to the bootstrapping protocol are demonstrated in Table~\ref{gps_results_table}.
\begin{table}[]
\caption{GNSS results on the test data set for each pipeline configuration. The precision, recall and F$_1$ scores are calculated on a macro basis. The asterisk indicates that this particular pipeline configuration gives statistically higher results}\label{gps_results_table}
\begin{center}
\scalebox{0.8}{
\begin{tabular}{cc|cccc}
\multicolumn{1}{l}{\textbf{}} & \multicolumn{1}{l|}{} & \multicolumn{4}{c}{\textbf{Metric}} \\ \hline
\multicolumn{1}{c|}{\textbf{Model}} & \textbf{Noise Removal} & \multicolumn{1}{P{1cm}|}{precision} & \multicolumn{1}{P{1cm}|}{recall} & \multicolumn{1}{P{1cm}|}{ F$_1$} & \multicolumn{1}{P{1cm}}{MCC} \\ \hline
\multicolumn{1}{c|}{\multirow{3}{*}{DT}} & no noise removal & \multicolumn{1}{c|}{48.52$\pm$3.34} & \multicolumn{1}{c|}{46.38$\pm$1.86} & \multicolumn{1}{c|}{45.08$\pm$2.24} & 27.08$\pm$3.14 \\ \cline{2-6} 
\multicolumn{1}{c|}{} & noise removal on raw location data & \multicolumn{1}{c|}{57.26$\pm$11.91} & \multicolumn{1}{c|}{56.48$\pm$10.55} & \multicolumn{1}{c|}{54.64$\pm$11.18} & 40.94$\pm$14.39 \\ \cline{2-6} 
\multicolumn{1}{c|}{} & noise removal on features & \multicolumn{1}{c|}{56.37$\pm$11.38} & \multicolumn{1}{c|}{55.58$\pm$9.03} & \multicolumn{1}{c|}{54.33$\pm$9.60} & 39.61$\pm$11.82 \\ \hline
\multicolumn{1}{c|}{\multirow{3}{*}{RF}} & no noise removal & \multicolumn{1}{c|}{46.89$\pm$5.08} & \multicolumn{1}{c|}{47.30$\pm$4.47} & \multicolumn{1}{c|}{44.10$\pm$5.94} & 28.83$\pm$5.72 \\ \cline{2-6} 
\multicolumn{1}{c|}{} & noise removal on raw location data & \multicolumn{1}{c|}{\textbf{62.04$\pm$4.19*}} & \multicolumn{1}{c|}{\textbf{66.88$\pm$4.36*}} & \multicolumn{1}{c|}{\textbf{62.31$\pm$4.18*}} & \textbf{50.13$\pm$5.73*} \\ \cline{2-6} 
\multicolumn{1}{c|}{} & \multicolumn{1}{l|}{noise removal on features} & \multicolumn{1}{c|}{58.74$\pm$8.83} & \multicolumn{1}{c|}{60.06$\pm$8.00} & \multicolumn{1}{c|}{57.47$\pm$7.79} & 44.75$\pm$10.60 \\ \hline
\multicolumn{1}{c|}{\multirow{3}{*}{SVM}} & no noise removal & \multicolumn{1}{c|}{46.10$\pm$4.14} & \multicolumn{1}{c|}{47.03$\pm$3.84} & \multicolumn{1}{c|}{44.33$\pm$4.61} & 27.46$\pm$5.36 \\ \cline{2-6} 
\multicolumn{1}{c|}{} & noise removal on raw location data & \multicolumn{1}{c|}{59.10$\pm$7.81} & \multicolumn{1}{c|}{59.21$\pm$4.94} & \multicolumn{1}{c|}{57.25$\pm$6.12} & 44.89$\pm$7.00 \\ \cline{2-6} 
\multicolumn{1}{c|}{} & noise removal on features & \multicolumn{1}{c|}{55.96$\pm$7.15} & \multicolumn{1}{c|}{55.14$\pm$6.31} & \multicolumn{1}{c|}{53.63$\pm$7.26} & 38.46$\pm$8.04 
\end{tabular}
}
\end{center}
\end{table}

We performed a statistical significance test with a significance level of 0.05 as described in the work of Wang et al.~\cite{wang2022towards} to determine if a certain pipeline configuration gave statistically higher performance than another one. First, we checked if the underlying distribution of each set of 50 performance scores was uniformly distributed. By performing an Anderson-Darling test~\cite{anderson1952asymptotic}, we concluded that the sets were not sampled from a Gaussian distribution. Given that the distributions were extracted from the same population, we decided to utilise the non-parametric Wilcoxon signed-rank test~\cite{wilcoxon1992individual}.

Based on Table~\ref{gps_results_table}, the pipeline configuration yielding noise removal on the raw location data in combination with an RF model gives statistically higher results for the precision, recall, F$_1$ and MCC scores than the other possible pipeline combinations. Thus, for the GNSS, the highest performance over all four score metrics was achieved with a pipeline configuration yielding noise removal on the raw location data in combination with an RF model. Additionally, the advantage of applying the Savitzky Golay filter on the raw location data or the extracted features in terms of acquiring a higher performance accuracy is reflected in the results demonstrated in Table~\ref{gps_results_table}. For all four score metrics, the lowest scores were obtained with pipelines yielding no noise removal application.
\subsubsection{UWB results:}\label{uwb_results}
Analogously, we assessed the 9 pipeline configurations on the collected UWB data. The Anderson-Darling test confirmed that the sets per pipeline configuration were not Gaussian distributed. For this reason, we used the Wilcoxon signed-rank test for the statistical analysis. The averaged test data set results over the 50 independent retrained models are reported in Table~\ref{uwb_results_table} in which bold-faced scores again indicate the highest performance and additional asterisks refer to the pipeline configuration that performs statistically better than other pipeline configurations per metric:
\begin{table}[]
\caption{UWB results on the test data set for each pipeline configuration. The precision, recall and F$_1$ scores are calculated on a macro basis.}\label{uwb_results_table}
\begin{center}
\scalebox{0.8}{
\begin{tabular}{cc|cccc}
\multicolumn{1}{l}{\textbf{}} & \multicolumn{1}{l|}{} & \multicolumn{4}{c}{\textbf{Metric}} \\ \hline
\multicolumn{1}{c|}{\textbf{Model}} & \textbf{Noise Removal} & \multicolumn{1}{P{2cm}|}{precision} & \multicolumn{1}{P{2cm}|}{recall} & \multicolumn{1}{P{2cm}|}{ F$_1$} & \multicolumn{1}{P{2cm}}{MCC} \\ \hline
\multicolumn{1}{c|}{\multirow{3}{*}{DT}} & no noise removal & \multicolumn{1}{c|}{80.28$\pm$5.19} & \multicolumn{1}{c|}{79.14$\pm$5.65} & \multicolumn{1}{c|}{78.87$\pm$5.83} & 71.90$\pm$7.49 \\ \cline{2-6} 
\multicolumn{1}{c|}{} & noise removal on raw location data & \multicolumn{1}{c|}{78.97$\pm$7.20} & \multicolumn{1}{c|}{76.79$\pm$8.01} & \multicolumn{1}{c|}{75.56$\pm$9.01} & 70.02$\pm$10.61 \\ \cline{2-6} 
\multicolumn{1}{c|}{} & noise removal on features & \multicolumn{1}{c|}{74.24$\pm$10.24} & \multicolumn{1}{c|}{69.22$\pm$13.04} & \multicolumn{1}{c|}{68.19$\pm$14.04} & 60.81$\pm$16.26 \\ \hline
\multicolumn{1}{c|}{\multirow{3}{*}{RF}} & no noise removal & \multicolumn{1}{c|}{86.78$\pm$1.95} & \multicolumn{1}{c|}{86.00$\pm$2.52} & \multicolumn{1}{c|}{85.87$\pm$2.35} & 81.39$\pm$3.01 \\ \cline{2-6} 
\multicolumn{1}{c|}{} & noise removal on raw location data & \multicolumn{1}{c|}{85.78$\pm$5.76} & \multicolumn{1}{c|}{\textbf{89.11$\pm$3.87}} & \multicolumn{1}{c|}{85.31$\pm$6.53} & 82.06$\pm$7.07 \\ \cline{2-6} 
\multicolumn{1}{c|}{} & \multicolumn{1}{l|}{noise removal on features} & \multicolumn{1}{c|}{83.15$\pm$5.54} & \multicolumn{1}{c|}{71.15$\pm$11.00} & \multicolumn{1}{c|}{68.72$\pm$13.18} & 67.73$\pm$11.74 \\ \hline
\multicolumn{1}{c|}{\multirow{3}{*}{SVM}} & no noise removal & \multicolumn{1}{c|}{89.74$\pm$1.64} & \multicolumn{1}{c|}{87.24$\pm$3.09} & \multicolumn{1}{c|}{\textbf{85.92$\pm$3.60}} & \textbf{82.85$\pm$3.91} \\ \cline{2-6} 
\multicolumn{1}{c|}{} & noise removal on raw location data & \multicolumn{1}{c|}{\textbf{89.95$\pm$3.54}} & \multicolumn{1}{c|}{86.21$\pm$6.73} & \multicolumn{1}{c|}{84.99$\pm$8.11} & 82.32$\pm$8.05 \\ \cline{2-6} 
\multicolumn{1}{c|}{} & noise removal on features & \multicolumn{1}{c|}{89.68$\pm$3.51} & \multicolumn{1}{c|}{83.17$\pm$7.83} & \multicolumn{1}{c|}{83.82$\pm$8.23} & 80.01$\pm$8.87 
\end{tabular}
}
\end{center}
\end{table}

Based on Table~\ref{uwb_results_table}, we can conclude that (i) noise removal on the raw location data with an SVM gives the highest results for the precision, (ii) noise removal on the raw location data with an RF gives the highest results for the recall, and (iii) no noise removal with an SVM gives the highest results for the F$_1$ and MCC. However, these particular pipeline configurations are not statistically the ones with the best performance per score metric. Therefore, we opted for the pipeline configuration that yields the highest results across the majority of score metrics, which is the configuration consisting of no noise removal in combination with an SVM.
\subsubsection{Comparison between GNSS and UWB:}
Regarding the GNSS, we concluded that the pipeline configuration containing noise removal on the raw location data with an RF model gave the highest performance results for all four performance metrics based on Table~\ref{gps_results_table}. As for the UWB, we concluded that the pipeline configuration consisting of no noise removal with an SVM model gave the highest performance results for the majority of performance metrics (i.e., F$_1$ and MCC) based on Table~\ref{uwb_results_table}. Therefore, we decided to compare the GNSS versus the UWB technology based on these respective pipeline configurations. Given that the sampled results per pipeline configuration are not uniformly distributed and we want to compare samples from two different populations, we opted for the non-parametric Mann-Whitney U test~\cite{mann1947test} with a standard confidence level of 95\%. 

For the precision, we observed a positive effect for the UWB with a significance level of 99\%. As for the recall, an identical positive effect for the UWB could be observed yielding a significance level of 99\%. Regarding the F$_1$ score, a positive effect could be observed in favour of the UWB with a significance level of 99\%. Likewise, for the MCC metric, a positive effect with a significance level of 99\% was present for the UWB. Thus, the UWB pipeline statistically outperforms the GNSS pipeline across all four score metrics.
\section{Conclusion and further research}\label{conclusions}
In this study, we designed and evaluated an automated framework to classify pre-defined human movement patterns collected with GNSS and UWB technology. The automated framework takes raw spatio-temporal data from both the GNSS and UWB systems, pre-processes this raw data to compute segments as well as global and local features, and at the end classifies the data into the pre-defined trajectory movement patterns. The automated framework allows for splitting the original trajectory into segments and applies three options for noise removal: (i) no noise removal, (ii) Savitzky Golay filter on the raw location data or (iii) Savitzky Golay filter on the extracted features, and three options for classification algorithms: (i) DT, (ii) RF or (iii) SVM. Additionally, we collected a total of 104 5-minute-long discrete GNSS and similar discrete UWB trajectories consisting of 4 unique movement patterns. Based on these collected discrete trajectories, we tested nine different pipeline configurations of our automated framework via a bootstrapping protocol to assess which one yielded the best performance. The hyperparameter optimalisation was automated by the SMAC method.

For the GNSS technology, we achieved the highest performance with a pipeline consisting of noise removal applied on the raw location data in combination with an RF model. Regarding the UWB technology, the pipeline with no noise removal in combination with an SVM model gave the highest performance for the F$_1$ and MCC score metrics. We compared these two pipeline configurations and concluded, based on a significance level of 99\% , that UWB achieved significantly better than GPS in classifying the pre-defined movement patterns in areas of approximately 100 square meters.

The movement patterns were limited to only four basic pre-defined movement patterns in our experiments. In future work, we could include more data with various complex movement patterns to make the pipeline configuration more applicable to real-life complex human movement behaviour. Thereby, a larger amount of trajectory data collections could enable future studies to investigate the use of more advanced classification models based on Recurrent Neural Networks (RNN). Local trajectory shape recognition could also be the solution to identify more complex trajectories. Here, sequence labelling can play a role to learn algorithms capable of classifying sequential patterns such as LSTM's, to recognize typical behaviour of humans based on long-lasting complex trajectories. Furthermore, in real-world scenarios, movement pattern data might not always include the ground truth. Thus, further research might explore the use of unsupervised classification algorithms~\cite{yao2018learning}, instead of supervised algorithms, to cluster trajectories in various patterns. Finally, the usage of additional sensors such as an accelerometer and a gyroscope might improve the performance of GNSS and UWB systems.
\bibliographystyle{splncs04}
\bibliography{bibliography}

\begin{thebibliography}{10}
\providecommand{\url}[1]{\texttt{#1}}
\providecommand{\urlprefix}{URL }
\providecommand{\doi}[1]{https://doi.org/#1}

\bibitem{anderson1952asymptotic}
Anderson, T.W., Darling, D.A.: Asymptotic theory of certain" goodness of fit"
  criteria based on stochastic processes. The annals of mathematical statistics
  pp. 193--212 (1952)

\bibitem{baratchi2013recognition}
Baratchi, M., Meratnia, N., Havinga, P.J.: Recognition of periodic behavioral
  patterns from streaming mobility data. In: International Conference on Mobile
  and Ubiquitous Systems: Computing, Networking, and Services. pp. 102--115.
  Springer (2013)

\bibitem{barrios2011improving}
Barrios, C., Motai, Y.: Improving estimation of vehicle's trajectory using the
  latest global positioning system with kalman filtering. IEEE Transactions on
  Instrumentation and Measurement  \textbf{60}(12),  3747--3755 (2011)

\bibitem{bastida2018accuracy}
Bastida~Castillo, A., G{\'o}mez~Carmona, C.D., De~la Cruz~S{\'a}nchez, E.,
  Pino~Ortega, J.: Accuracy, intra-and inter-unit reliability, and comparison
  between gps and uwb-based position-tracking systems used for time--motion
  analyses in soccer. European journal of sport science  \textbf{18}(4),
  450--457 (2018)

\bibitem{chicco2020advantages}
Chicco, D., Jurman, G.: The advantages of the matthews correlation coefficient
  (mcc) over f1 score and accuracy in binary classification evaluation. BMC
  genomics  \textbf{21}(1),  1--13 (2020)

\bibitem{dabiri2018inferring}
Dabiri, S., Heaslip, K.: Inferring transportation modes from gps trajectories
  using a convolutional neural network. Transportation research part C:
  emerging technologies  \textbf{86},  360--371 (2018)

\bibitem{dabove2019towards}
Dabove, P., Di~Pietra, V.: Towards high accuracy gnss real-time positioning
  with smartphones. Advances in Space Research  \textbf{63}(1),  94--102 (2019)

\bibitem{etemad2018predicting}
Etemad, M., Soares~J{\'u}nior, A., Matwin, S.: Predicting transportation modes
  of gps trajectories using feature engineering and noise removal. In: Canadian
  conference on artificial intelligence. pp. 259--264. Springer (2018)

\bibitem{feurer_hyperparameter_2019}
Feurer, M., Hutter, F.: Hyperparameter optimization. pp. 3--38. Springer (2019)

\bibitem{gorodkin2004comparing}
Gorodkin, J.: Comparing two k-category assignments by a k-category correlation
  coefficient. Computational biology and chemistry  \textbf{28}(5-6),  367--374
  (2004)

\bibitem{grandini2020metrics}
Grandini, M., Bagli, E., Visani, G.: Metrics for multi-class classification: an
  overview. arXiv preprint arXiv:2008.05756  (2020)

\bibitem{han2011data}
Han, J., Pei, J., Kamber, M.: Data mining: concepts and techniques. Elsevier
  (2011)

\bibitem{kalman1960new}
Kalman, R.E.: A new approach to linear filtering and prediction problems
  (1960)

\bibitem{kaplan2017understanding}
Kaplan, E.D., Hegarty, C.: Understanding GPS/GNSS: principles and applications.
  Artech house (2017)

\bibitem{kennedy2020improving}
Kennedy, H.L.: Improving the frequency response of savitzky-golay filters via
  colored-noise models. Digital Signal Processing  \textbf{102},  102743 (2020)

\bibitem{kuhn2008high}
Kuhn, M., Zhang, C., Merkl, B., Yang, D., Wang, Y., Mahfouz, M., Fathy, A.:
  High accuracy uwb localization in dense indoor environments. In: 2008 IEEE
  International Conference on Ultra-Wideband. vol.~2, pp. 129--132. IEEE (2008)

\bibitem{laanen2022classification}
Laanen, R.: Classification of pre-defined movement patterns: A comparison
  between gnss and uwb technology. LIACS  (2022)

\bibitem{lin2018gps}
Lin, Q., Liu, X., Wang, W.: Gps trajectories based personalized safe geofence
  for elders with dementia. In: 2018 IEEE SmartWorld, Ubiquitous Intelligence
  \& Computing, Advanced \& Trusted Computing, Scalable Computing \&
  Communications, Cloud \& Big Data Computing, Internet of People and Smart
  City Innovation (SmartWorld/SCALCOM/UIC/ATC/CBDCom/IOP/SCI). pp. 505--514.
  IEEE (2018)

\bibitem{lindauer-arxiv21a}
Lindauer, M., Eggensperger, K., Feurer, M., Biedenkapp, A., Deng, D.,
  Benjamins, C., Ruhkopf, T., Sass, R., Hutter, F.: Smac3: A versatile bayesian
  optimization package for hyperparameter optimization. In: ArXiv: 2109.09831
  (2021), \url{https://arxiv.org/abs/2109.09831}

\bibitem{luo2020outdoor}
Luo, Y., Irvin, D., Buss, N., Gutierrez, A., Rous, B.: Outdoor playground
  localization system for tracking young children using ubisense sensor
  network. In: 2020 IEEE 17th International Conference on Mobile Ad Hoc and
  Sensor Systems (MASS). pp. 7--12. IEEE (2020)

\bibitem{mahfouz2008investigation}
Mahfouz, M.R., Zhang, C., Merkl, B.C., Kuhn, M.J., Fathy, A.E.: Investigation
  of high-accuracy indoor 3-d positioning using uwb technology. IEEE
  Transactions on Microwave Theory and Techniques  \textbf{56}(6),  1316--1330
  (2008)

\bibitem{mann1947test}
Mann, H.B., Whitney, D.R.: On a test of whether one of two random variables is
  stochastically larger than the other. The annals of mathematical statistics
  pp. 50--60 (1947)

\bibitem{marins2013extending}
Marins, F., Rodrigues, R., Portela, F., Santos, M., Abelha, A., Machado, J.:
  Extending a patient monitoring system with identification and localisation.
  In: 2013 IEEE International Conference on Industrial Engineering and
  Engineering Management. pp. 1082--1086. IEEE (2013)

\bibitem{mimoune2019evaluation}
Mimoune, K.M., Ahriz, I., Guillory, J.: Evaluation and improvement of
  localization algorithms based on uwb pozyx system. In: 2019 International
  Conference on Software, Telecommunications and Computer Networks (SoftCOM).
  pp.~1--5. IEEE (2019)

\bibitem{nasri2022novel}
Nasri, M., Tsou, Y.T., Koutamanis, A., Baratchi, M., Giest, S., Reidsma, D.,
  Rieffe, C.: A novel data-driven approach to examine children’s movements
  and social behaviour in schoolyard environments. Children  \textbf{9}(8),
  ~1177 (2022)

\bibitem{scikit-learn}
Pedregosa, F., Varoquaux, G., Gramfort, A., Michel, V., Thirion, B., Grisel,
  O., Blondel, M., Prettenhofer, P., Weiss, R., Dubourg, V., Vanderplas, J.,
  Passos, A., Cournapeau, D., Brucher, M., Perrot, M., Duchesnay, E.:
  Scikit-learn: Machine learning in {P}ython. Journal of Machine Learning
  Research  \textbf{12},  2825--2830 (2011)

\bibitem{PETRE2017309}
Petre, A.C., Chilipirea, C., Baratchi, M., Dobre, C., {van Steen}, M.: Chapter
  14 - wifi tracking of pedestrian behavior. In: Xhafa, F., Leu, F.Y., Hung,
  L.L. (eds.) Smart Sensors Networks, pp. 309--337. Intelligent Data-Centric
  Systems, Academic Press (2017).
  \doi{https://doi.org/10.1016/B978-0-12-809859-2.00018-8},
  \url{https://www.sciencedirect.com/science/article/pii/B9780128098592000188}

\bibitem{pozyx}
Pozyx: Pozyx. \url{https://www.pozyx.io} (2022)

\bibitem{pozyxcreatorkit}
Pozyx: Pozyx creator kit. \url{https://www.pozyx.io/creator} (2022)

\bibitem{TabA7}
Samsung: Samsung galaxy tab a7 lite.
  \url{https://www.samsung.com/nl/tablets/galaxy-tab-a/galaxy-tab-a7-lite-lte-gray-32gb-sm-t225nzaaeub/}
  (2022)

\bibitem{savitzky1964smoothing}
Savitzky, A., Golay, M.J.: Smoothing and differentiation of data by simplified
  least squares procedures. Analytical chemistry  \textbf{36}(8),  1627--1639
  (1964)

\bibitem{schafer2011savitzky}
Schafer, R.W.: What is a savitzky-golay filter?[lecture notes]. IEEE Signal
  processing magazine  \textbf{28}(4),  111--117 (2011)

\bibitem{shih2016personal}
Shih, D.H., Shih, M.H., Yen, D.C., Hsu, J.H.: Personal mobility pattern mining
  and anomaly detection in the gps era. Cartography and Geographic Information
  Science  \textbf{43}(1),  55--67 (2016)

\bibitem{sokolova2009systematic}
Sokolova, M., Lapalme, G.: A systematic analysis of performance measures for
  classification tasks. Information processing \& management  \textbf{45}(4),
  427--437 (2009)

\bibitem{tomavstik2017horizontal}
Toma{\v{s}}t{\'\i}k~Jr, J., Toma{\v{s}}t{\'\i}k~Sr, J., Salo{\v{n}}, {\v{S}}.,
  Piroh, R.: Horizontal accuracy and applicability of smartphone gnss
  positioning in forests. Forestry: An International Journal of Forest Research
   \textbf{90}(2),  187--198 (2017)

\bibitem{wang2019automated}
Wang, C., B{\"a}ck, T., Hoos, H.H., Baratchi, M., Limmer, S., Olhofer, M.:
  Automated machine learning for short-term electric load forecasting. In: 2019
  IEEE Symposium Series on Computational Intelligence (SSCI). pp. 314--321.
  IEEE (2019)

\bibitem{wang2022towards}
Wang, C., Baratchi, M., B{\"a}ck, T., Hoos, H.H., Limmer, S., Markus, O.:
  Towards time-series-specific feature engineering in automated machine
  learning frameworks for time-series forecasting. IEEE  (under review)

\bibitem{waqar2021analysis}
Waqar, A., Ahmad, I., Habibi, D., Phung, Q.V.: Analysis of gps and uwb
  positioning system for athlete tracking. Measurement: Sensors  \textbf{14},
  100036 (2021)

\bibitem{wilcoxon1992individual}
Wilcoxon, F.: Individual comparisons by ranking methods. In: Breakthroughs in
  statistics, pp. 196--202. Springer (1992)

\bibitem{yao2018learning}
Yao, D., Zhang, C., Zhu, Z., Hu, Q., Wang, Z., Huang, J., Bi, J.: Learning deep
  representation for trajectory clustering. Expert Systems  \textbf{35}(2),
  e12252 (2018)

\end{thebibliography}

\end{document}